\documentclass[conference]{IEEEtran}
\usepackage{times}
\usepackage{graphicx}
\usepackage[numbers]{natbib}
\usepackage{multicol}
\usepackage[table]{xcolor}
\usepackage[
    bookmarks=true,
    colorlinks=true,
    allcolors=blue
]{hyperref}
\usepackage{lipsum} 
\usepackage{listings}
\definecolor{pygKeyword}{HTML}{008000}
\definecolor{pygString}{HTML}{BA2121}
\definecolor{pygComment}{HTML}{408080}
\definecolor{pygBuiltin}{HTML}{008000}
\definecolor{pygNumber}{HTML}{666666}
\definecolor{pygFunc}{HTML}{0000FF}
\lstdefinestyle{pythonstyle}{
    language=Python,
    basicstyle=\footnotesize\ttfamily,
    keywordstyle=\bfseries\color{pygKeyword},
    commentstyle=\itshape\color{pygComment},
    stringstyle=\color{pygString},
    numberstyle=\color{pygNumber},
    emphstyle=\color{pygFunc},
    emph={[2]self,cls},
    emphstyle={[2]\color{pygBuiltin}},
    showstringspaces=false,
    breaklines=true,
    breakatwhitespace=false,
    frame=single,
    framesep=2mm,
    xleftmargin=0pt,
    tabsize=2,
    columns=fullflexible,
    keepspaces=true,
}
\usepackage{cleveref}
\usepackage{xspace}

\usepackage{comment}
\usepackage{booktabs}
\usepackage{cleveref}
\usepackage{array}
\usepackage{caption}
\usepackage{float}
\usepackage{multicol}
\usepackage{tablefootnote}
\usepackage{amssymb}
\usepackage{placeins}
\usepackage{makecell}

\makeatletter
\renewcommand\hyper@natlinkbreak[2]{#1}
\makeatother

\newcommand{\pagelink}{robot-i-o.github.io}

\newcommand{\proj}{Robot I/O\xspace}
\newcommand{\PROJ}{RIO\xspace}

\usepackage{twemojis}
\newcommand{\Y}{\twemoji{check mark}}
\newcommand{\N}{\twemoji{cross mark}}  %

\usepackage{authblk}
\renewcommand{\footnoterule}{\kern-3pt \hrule width 0.4\columnwidth \kern 2.6pt}
\usepackage{comment}
\usepackage{dblfloatfix}
\begin{document}

\title{RIO: Flexible Real-Time Robot I/O for Cross-Embodiment Robot Learning}

\newcommand\blfootnote[1]{%
  \begingroup
  \renewcommand\thefootnote{}\footnote{#1}%
  \addtocounter{footnote}{-1}%
  \endgroup
}
\makeatletter
\renewcommand\AB@affilsepx{   \protect\Affilfont}
\makeatother

\author[1]{Pablo Ortega-Kral$^{*}$}
\author[1]{Eliot Xing$^{*}$}
\author[1]{Arthur Fender Coelho Bucker}
\author[1]{Vernon Luk}
\author[1,2]{Junseo Kim}
\author[1]{\authorcr Owen Kwon}
\author[1]{Angchen Xie}
\author[1]{Nikhil Sobanbabu}
\author[1]{Yifu Yuan}
\author[1,4]{Megan Lee}
\author[1]{Deepam Ameria}
\author[1]{\authorcr Bhaswanth Ayapilla}
\author[3]{Jaycie Bussell}
\author[1]{Guanya Shi}
\author[1,3]{Jonathan Francis}
\author[1,4]{Jean Oh$^{\dagger}$}
\affil[1]{Carnegie Mellon University   }
\affil[2]{TU Delft   }
\affil[3]{Bosch Center for AI   }
\affil[4]{Lavoro AI}

\makeatletter
\let\old@maketitle\@maketitle
\renewcommand{\@maketitle}{
  \begin{center}
  \old@maketitle

    \begin{minipage}{0.35\textwidth}
        \centering
        \includegraphics[width=\linewidth,trim={6cm 5cm 6cm 0},clip]{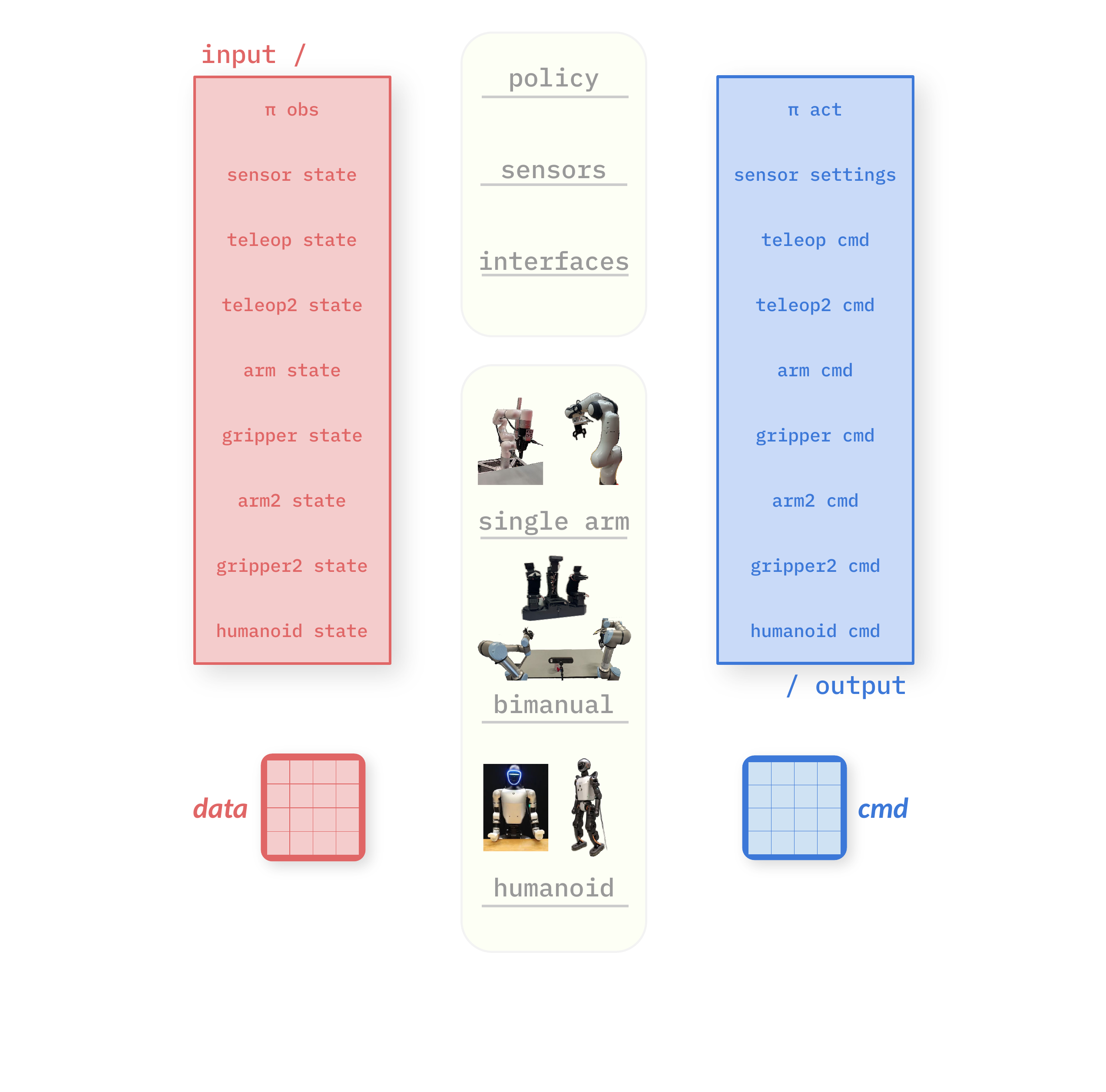}
    \end{minipage}
    \hfill
    \begin{minipage}{0.63\textwidth}
        \centering
        \includegraphics[width=\linewidth]{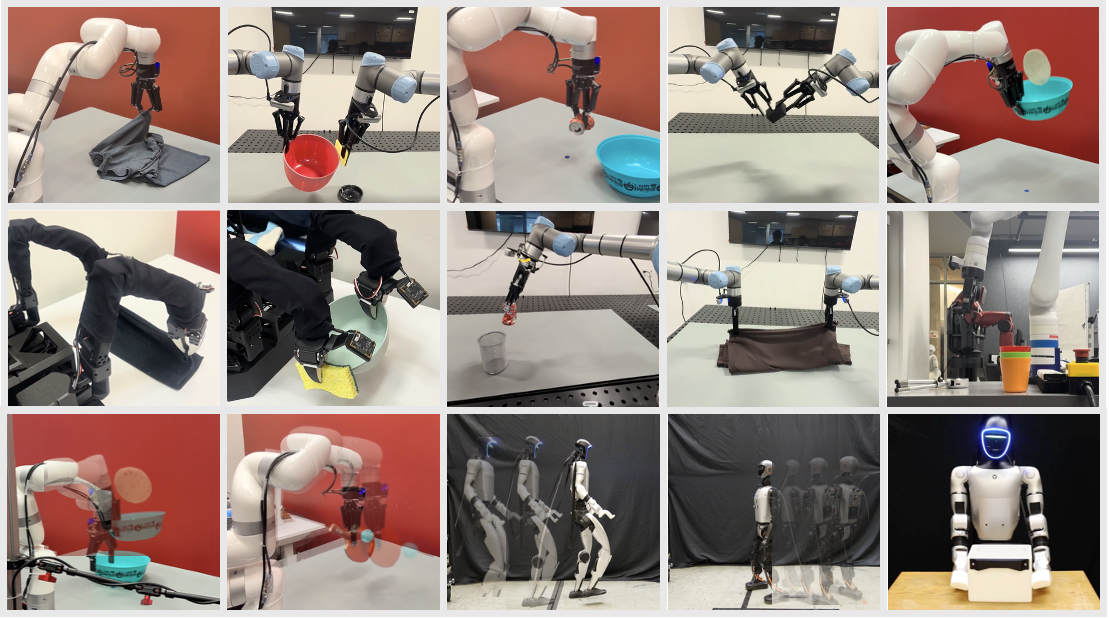}
    \end{minipage}
    \captionof{figure}{\textbf{System overview.} We introduce \PROJ, flexible real-time \proj for cross-embodiment robot learning, a lightweight Python-based framework to coordinate diverse robot morphologies, sensors, teleoperation interfaces, and policies.}
    \label{fig:overview}
    \vspace{-1.5em}
  \end{center}
}
\makeatother
\let\oldmaketitle\maketitle
\renewcommand{\maketitle}{
  \oldmaketitle
  \addtocounter{figure}{-1}
}

\maketitle
\blfootnote{$^{*}$Equal contribution, $^{\dagger}$Corresponding author: \texttt{jeanoh@cmu.edu}}
\begin{abstract}
Despite recent efforts to collect multi-task, multi-embodiment datasets, to design recipes for training Vision-Language-Action models (VLAs), and to showcase these models on different robot platforms, generalist cross-embodiment robot capabilities remains a largely elusive ideal. Progress is limited by fragmented infrastructure: most robot code is highly specific to the exact setup the user decided on, which adds major overhead when attempting to reuse, recycle, or share artifacts between users.
We present \PROJ (\proj), an open source Python framework that provides flexible, lightweight components for robot control, teleoperation, data formatting, sensor configuration, and policy deployment across diverse hardware platforms and morphologies. \PROJ provides abstractions that enable users to make any choice and to switch between them, with minimal reconfiguration effort.
We validate \PROJ on VLA deployment workflows across three morphologies (single-arm, bimanual, humanoid) and four hardware platforms with varying grippers and cameras. Using teleoperated data collected with \PROJ, we fine-tune state-of-the-art VLAs including $\pi_{0.5}$ and GR00T on household tasks such as pick-and-place, folding, and bowl scrubbing. By open sourcing all our efforts, we hope the community can accelerate their pace of robot learning on real-world robot hardware.
Additional details at:~\href{https://robot-i-o.github.io/}{\texttt{\pagelink}}.
\end{abstract}

\IEEEpeerreviewmaketitle

\section{Introduction}

Vision-language-action models (VLAs) have recently emerged as a promising approach for training generalist robot policies, leveraging large-scale cross-embodiment datasets to learn broadly capable robot behaviors. Despite their potential, successfully deploying VLAs on new robot morphologies and platforms remains challenging in practice. VLAs are difficult to run out-of-the-box on new embodiments, and getting them to work still demands substantial engineering effort. This difficulty, however, extends well beyond VLAs.

Robotics practitioners have long contended with the fragmentation inherent in the field. Varying morphologies, diverse sensor configurations, heterogeneous hardware platforms, and manufacturer-specific driver code collectively result in robot infrastructure that is highly specific to a user's particular setup. This leads to significant overhead when attempting to reuse code, share datasets, or build on each other's work. Existing cross-embodiment datasets like Open X-Embodiment~\cite{collaboration_open_2025} are, in practice, aggregations of many individual collection efforts conducted across disparate infrastructure.

The cost of this fragmentation is growing. As robot hardware becomes increasingly affordable, more platforms are entering research labs and deployment settings. Yet, the specialized nature of most robotics infrastructure means that each new platform carries substantial integration overhead. Consider a common scenario: a research team wishes to reproduce real-world results released by another group. To use the original control code, they would need to replicate the exact hardware setup one-to-one, as in efforts like DROID~\cite{khazatsky_droid_2025}. If they instead have a different robot arm, then they face the burden of rewriting the entire control stack from scratch, before even trying to adapt any learned policies. This makes most robot learning hardware code \textit{difficult to reuse}, and switching between platforms far harder than it should be. 

What is the most important infrastructure for robot learning to advance? Beyond large datasets, we believe that a lack of flexible, reusable, accessible, and performant full-stack robot infrastructure has been a critical barrier to cumulative progress and collaboration within the field. Robot learning is missing reusable building blocks for hardware with flexible abstractions that have become standard in other areas of machine learning. Just as specialized high-performance GPU kernels within high-level auto-differentiation frameworks have enabled the rapid development and iteration of neural networks, robotics requires analogous foundational components for hardware and control that can be reliably shared, extended, and built upon across the community.

In this paper, we present the following contributions:
\begin{itemize}
    \item[\textbf{i)}] We introduce \PROJ, a flexible real-time \proj framework that provides reusable infrastructure for data collection and policy deployment across diverse robot embodiments. \PROJ is not intended as a comprehensive robot learning solution, but rather a lightweight set of reusable building blocks that can be quickly combined to deploy policies on real robots, tailored to each user's desired configuration. \PROJ is designed to be flexible, reusable, accessible, performant, and consistent, with abstractions such that the user is free to make any choice at each layer of the stack and switch between them with minimal effort.
    \item[\textbf{ii)}] We validate \PROJ for the VLA deployment workflow spanning diverse embodiments across single arm, bimanual, and humanoid robots with different grippers and sensors. This includes different robots, cameras, teleoperation interfaces, middlewares, data formats, and policies.
    \item[\textbf{iii)}] We demonstrate real-world deployment by collecting teleoperated data to fine-tune state-of-the-art VLAs such as $\pi_{0.5}$ and GR00T, on household tasks such as pick-and-place, folding, and bowl scrubbing, including a cross-embodiment policy trained on mixed data from two different robot morphologies and platforms.
\end{itemize}

\section{Related Works}

\begin{table*}[t]
\centering
\footnotesize
{\setlength{\tabcolsep}{3pt}
\caption{\textbf{Comparison of cross-embodiment robot stacks.} We compare various cross-embodiment robot stacks on the basis of native platform support, at different layers of the robot learning pipeline, e.g., data-collection system support, robot hardware support, middleware, data formats, and policy architecture support. For example, some stacks combine robot arm and robot gripper drivers, making it difficult to use other end effectors on arms.}
\label{tab:stacks}
\resizebox{\textwidth}{!}{
\rowcolors{2}{white}{lightgray!15}
\begin{tabular}{l c c c c c c l l l}
\toprule
Framework & Humanoids & Bimanual & Single arm & Robot grippers & Teleop & Cameras & Middleware(s) & Data format(s) & Policies \\
\midrule
Ark~\cite{dierking2025ark} & \Y &\Y& \Y & \N & \Y & \Y & \N \::\: LCM & \N \::\: Pickle & \Y \\
LeRobot~\cite{cadene2024lerobot} & \Y & \Y & \Y & \N & \Y & \Y & \N \::\: Threads/gRPC$^{1}$ & \N \::\: LeRobotDataset & \Y \\
ManiUniCon~\cite{maniunicon2025} & \N & \N & \Y & \N & \Y & \Y & \N \::\: Shm & \N \::\: Zarr & \Y \\
PAPRLE~\cite{kwon2025paprle} & \Y & \Y & \Y & \Y & \Y & \Y & \N \::\: ROS & \N \::\: Pickle & n/a \\
PyRobot~\cite{murali2019pyrobot} & \N & \N & \Y & \Y & \Y & \Y & \N \::\: ROS & \N \::\: Pickle & n/a \\
RCS~\cite{julg2025robot} & \N & \N & \Y & \Y & \Y & \Y & \N \::\: RPC & \N \::\: Parquet & \Y \\
RoBits~\cite{grotz2024peract2} & \N & \Y & \Y & \Y & \Y & \Y & \N \::\: ZMQ & \N \::\: NPZ/JSON & n/a \\
UMI, DP~\cite{chi2024universal, chi2025diffusion} & \N &\Y& \Y & \N & \Y & \Y & \N \::\: Shm & \N \::\: Zarr & \N \::\: DP \\
\textbf{\PROJ (ours)} & \Y &\Y& \Y & \Y & \Y & \Y & \Y \:: \:\textbf{any} & \Y \:: \:\textbf{any} & \Y \\
\bottomrule
\multicolumn{10}{l}{\rule{0pt}{2.5ex}\textit{$^{1}$LeRobot uses Threads for hardware drivers and gRPC for asynchronous policy inference.}}
\end{tabular}}
}
\vspace{-1.0em}
\end{table*}

\subsection{Generalist Robot Policies}
Recent advances in vision-language-action models (VLAs)~\cite{brohan2022rt, ghosh_octo_2024, black2024pi_0, intelligence2025pi_, team2025gemini, bjorck2025gr00t} aim to leverage the robust image-to-language alignment learned by internet-scale pre-trained vision-language models (VLMs)~\cite{zhai2023sigmoid, driess2023palm, beyer2024paligemma, bai2025qwen2, deitke2025molmo} to train generalist robot policies. VLAs adapt VLMs to predict actions through imitation learning on robot datasets collected via human teleoperation of robots, scaling foundational work on imitation learning for visuomotor policy learning, such as ALOHA~\cite{zhao_learning_2023} and Diffusion Policy~\cite{chi2025diffusion}. Due to the computational resources and data scale required, state-of-the-art VLAs are predominantly trained by industry labs with substantial infrastructure and engineering personnel. Open source efforts have sought to reproduce and democratize these results~\cite{wen2025tinyvla, liu2025rdt, starvla2025, deitke2025molmo, kim2025openvla}, providing fully open source implementations and model weights. However, a significant limitation remains: current VLAs must in practice be fine-tuned for each robot setup. Released VLA model checkpoints are typically fine-tuned for specific embodiments, such as the Franka arm from DROID~\cite{khazatsky_droid_2025} or the WidowX arm from BridgeData V2~\cite{walke2023bridgedata}. Consequently, end-users must either reproduce the exact hardware setup used during training~\cite{pi0-experiment-wild}, or undertake substantial engineering effort to implement their own robot control stack, before even attempting to adapt learned policies to their own platforms. In this work, we lower this barrier by introducing a flexible cross-embodiment robot control stack, validated on the VLA adaptation workflow and deployed across diverse robot configurations.

\subsection{Cross-Embodiment Robot Data}
The effectiveness of scaling VLAs depends on access to large-scale robot demonstration data. Prior work has established that scaling robot data across both task diversity and robot embodiments~\cite{dasari2020robonet, wu2024robomind, chen_robotwin_2025, doshi2025scaling, rdt2, wu2026pragmatic} shows promise at training better generalist robot policies, and cross-embodiment robot data may also enable learning directly from humans~\cite{kareer2025emergence}. Training truly general robot policies requires diversity in both tasks and embodiments. Open X-Embodiment~\cite{collaboration_open_2025} aggregates 60 datasets spanning over 1 million robot trajectories across 22 embodiments. However, the heterogeneous collection techniques and sensor configurations across these datasets necessitate substantial curation for effective policy training, such as through filtering~\cite{ghosh_octo_2024, khazatsky_droid_2025} or data mixture re-weighting~\cite{hejna2024remixoptimizingdatamixtures}. In this work, we aim to facilitate the collection of high-quality cross-embodiment robot data by developing flexible and reusable robot infrastructure.

\subsection{Robot Control Stacks}

Over the years, many robot control stacks have emerged~\cite{metta2006yarp, kumar2023robohive, dierking2025ark, cadene2024lerobot, maniunicon2025, kwon2025paprle, murali2019pyrobot, julg2025robot, grotz2024peract2, macenski2022robot, chi2024universal} that are capable of cross-embodiment robot control. ROS \cite{macenski2022robot} was developed to facilitate system compartmentalization and distributed communication. While this modular, distributed approach offers benefits for complex robotic systems, it requires compounding systems-level engineering to coordinate all modules together. Furthermore, ROS presents a high barrier to entry for researchers and practitioners new to robotics, as it requires wrangling a complex configuration and build system.

Despite this proliferation of frameworks, robot code remains highly platform-specific. We attribute this to two factors: first, most roboticists work with a single hardware platform, which incentivizes writing vendor-specific code quickly rather than abstractable solutions; second, existing frameworks lack flexible abstractions at every layer of the stack to ensure cross-embodiment robot code is easy to write in the first place. 
LeRobot~\cite{cadene2024lerobot} has also found mainstream adoption among the broader robotics community. Its popularity stems in part from its support for low-cost robot hardware along with Python-only implementations, which eliminates the need for complex build systems such as in ROS and multi-language dependencies. By streamlining the development workflow, LeRobot has lowered the barrier to deploying real-world robot systems and enabled a wider range of practitioners to experiment with robot learning. \PROJ shares the motivation to build an accessible open source community for cross-embodiment robot learning. 

In Table~\ref{tab:stacks}, we compare the flexibility of \PROJ to a variety of existing cross-embodiment robot infrastructure, across every layer of the robot learning stack. \PROJ offers a reusable Python-based set of real-time robot infrastructure in a similar style to DP/UMI and LeRobot, while supporting reconfigurability at each layer to facilitate VLA deployment workflows across diverse robot morphologies. \PROJ enables combinatorial configuration of robots, teleoperation devices, cameras, middlewares, data formats, and policies, for maximum flexibility.

\section{\PROJ (\proj)}

\PROJ is a Python-based framework for flexible real-time \proj, with reusable components for robot control, teleoperation, data collection, and policy deployment across diverse robot embodiments. Users are free to make any choices at every layer of the stack (humanoid robots, robot arms, robot grippers, teleoperation interfaces, cameras, middlewares, data formats, policies) and to switch between them with minimal effort for reconfigurability. See Table~\ref{tab:hardware} for a detailed list of currently supported hardware.

\begin{figure*}[t]
    \centering
    \includegraphics[width=0.8\textwidth,trim={0 1cm 0 3.5cm},clip]{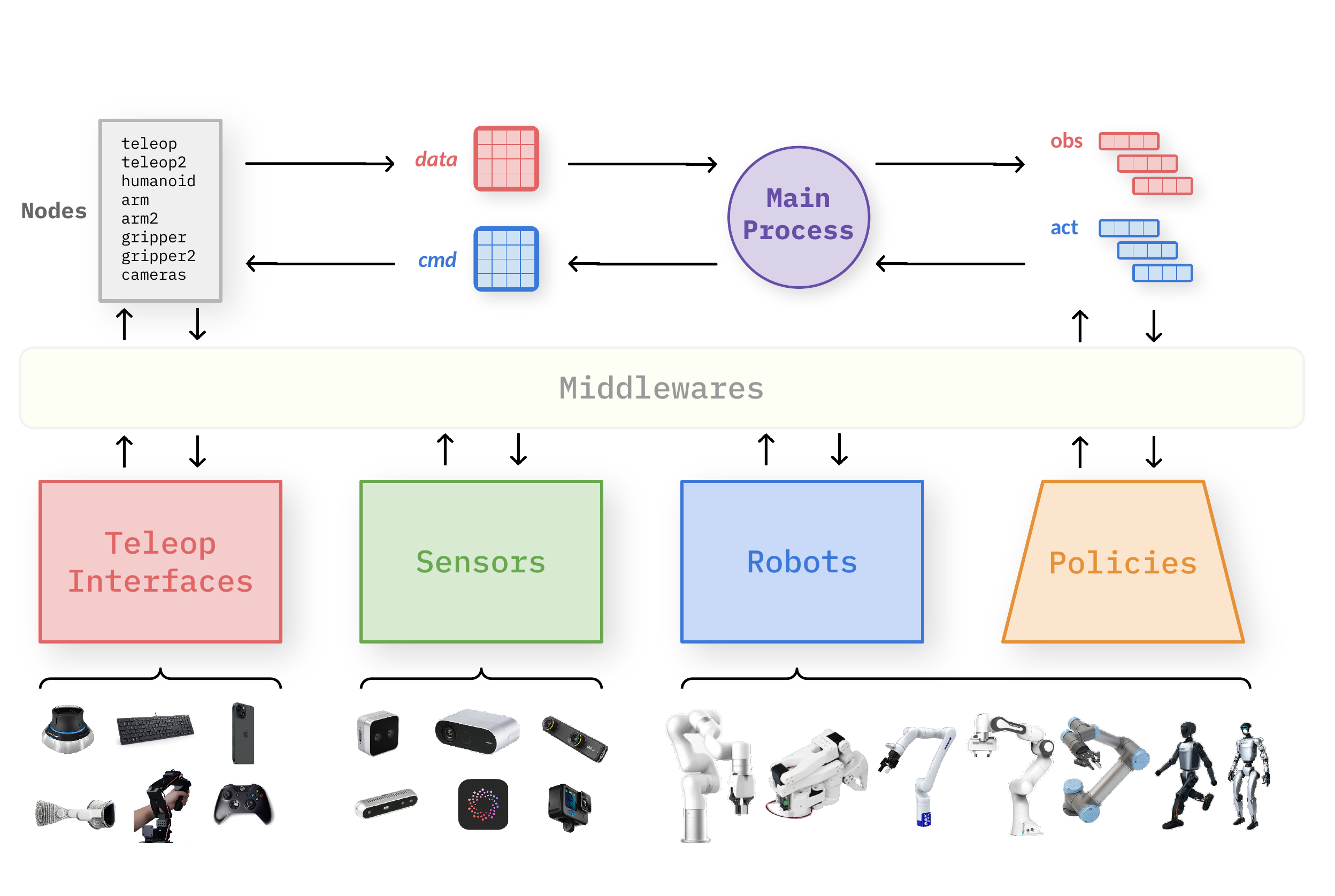}
    \caption{\textbf{Architecture.} High-level overview of the architecture of \PROJ. Every component of the stack is flexible, meaning that the user is free to choose between different options (robots, sensors, teleoperation interfaces, middlewares, data formats, policies) and switch between them, with minimal effort.}\vspace{-15pt}%
    \label{fig:architecture}%
\end{figure*}

Figure~\ref{fig:architecture} illustrates the overall system architecture of \PROJ. Section~\ref{sec:method_design} describes the \textit{Design Philosophy}, Section~\ref{sec:method_nodesmw} introduces the client-server \textit{Nodes} abstraction and \textit{Middlewares} implementation for message passing, Section~\ref{sec:method_station} outlines the reconfigurable instantiation of \textit{Robot Stations}, Section~\ref{sec:method_teleopdata} describes \textit{Teleoperation} and \textit{Data Collection} used to collect real-world robot trajectories across multiple embodiments, and Section~\ref{sec:method_policy} describes the implementation of asynchronous \textit{Policy Inference} to obtain smooth real-world rollouts with observation/action chunking.

\begin{table}[t]
\centering
\footnotesize
\caption{\textbf{Current hardware support} (\textit{more incoming}). \PROJ provides flexibility across robot hardware (humanoid robots, robot arms, robot grippers), teleop interfaces, cameras, and middlewares for multi-process distributed communication. These can be combined in any configuration depending on the user's hardware requirements.}
\label{tab:hardware}
\resizebox{\columnwidth}{!}{
\rowcolors{2}{white}{lightgray!15}
\begin{tabular}{@{}p{0.25\linewidth}>{\raggedright\arraybackslash}p{0.7\linewidth}@{}}
\toprule
\textbf{Humanoid Robots} & Unitree G1, Booster T1 \\
\textbf{Robot Arms} & UFactory (xArm5/6/7, 850, Lite6), UR (UR5e, UR7e), Franka (FR3, Panda), Kinova (Gen3), SO-100/SO-101 \\
\textbf{Robot Grippers} & UFactory Gripper, Franka Gripper, Robotiq Gripper (2F-85/2F-140), DH-Robotics Gripper (AG-105-145) \\
\textbf{Teleop Interfaces} & Spacemouse, Gamepad, Keyboard, VR (Apple Vision Pro, Meta Quest, Pico 4 Ultra), Leader-Follower (GELLO), Phone \\
\textbf{Cameras} & RealSense, ZED, UVC (Webcams, USB cameras), iPhone (Record3D) \\
\textbf{Middlewares} & Shared Memory, Thread, Portal, Zenoh, ZeroRpc \\
\bottomrule
\end{tabular}
}
\vspace{-5.0mm}
\end{table}

\subsection{Design Philosophy}
\label{sec:method_design}

We describe the five design tenets of \PROJ: \textit{flexible}, \textit{reusable}, \textit{accessible}, \textit{performant}, and \textit{consistent}.

\begin{itemize}
    \item \textit{Flexible.} \PROJ is agnostic to each component and does not make any locked-in choices. The user is free to choose between options at every layer, and switch between them with minimal effort.

    \item \textit{Reusable.} \PROJ is composed of a lightweight set of reusable building blocks, that can be quickly combined and modified to support a user's desired configuration. 

    \item \textit{Accessible.} \PROJ is quick to install and uses Python-based hardware modules, with a command-line interface over a single configuration file. 

    \item \textit{Performant.} \PROJ is fully capable of high-frequency real-time robot control, and uses asynchronous policy inference to yield smooth robot trajectories.

    \item \textit{Consistent.} \PROJ is designed to ensure consistent, scalable, reproducible data collection and robot learning. 

\end{itemize}

Taken together, our design values serve a single goal: abstract user logic should be written once and should be reusable on any robot station. Although a wide range of robot infrastructure exists today (Table~\ref{tab:stacks}), most robot code is highly platform-specific, with systems rarely reusable outside their original hardware setup. For a given task, such as teleoperation or policy deployment, \PROJ exposes abstract scripts that operate on arbitrary embodiments without modification. Swapping robot platforms, cameras, teleoperation devices, or middlewares is a change to the configuration rather than to the control logic. With this design, \PROJ aims to facilitate the writing of reusable control stack components rather than platform-specific code.

\subsection{Nodes and Middlewares}
\label{sec:method_nodesmw}

Nodes for teleoperation interfaces, sensors, robots, and policies are implemented from the same template Node, requiring minimal boilerplate to enable flexible, real-time I/O across diverse hardware and deployment configurations.

\noindent\textbf{Nodes.} A Node dynamically inherits from a given Middleware that automatically handles message passing. Factory functions produce matched server-client Node pairs, for which its dynamic parent class implements the specified Middleware. Nodes support three execution patterns for publishing data and handling requests between server-client pairs:
\begin{enumerate}
    \item Publish-only: \texttt{pub()} publishes data in the run loop.
    \item Request-only: \texttt{req()} handles requests in the run loop.
    \item Combined: \texttt{pubreq()} publishes data and handles requests in a single loop.
\end{enumerate}
For patterns (1) and (2), the complementary operation can be optionally run in a separate worker loop thread. For example, \texttt{pub()} in the run loop with \texttt{req()} in a worker loop thread, or vice versa. To implement a Node, the user defines the relevant methods: a \texttt{pub()} implementation calls \texttt{ring\_buffer.put(..)} to publish data, while a \texttt{req()} implementation calls \texttt{request\_queue.get()} to receive and process requests. Published data flows through a ring buffer that continuously streams state at a fixed frequency, providing time-synchronized access to sensor readings, robot poses, and other data. Requests flow through a queue that enables asynchronous command communication, allowing multiple clients to send timestamped commands independently and at arbitrary rates. For each, a server Node executes ``publish/request'', while a client Node automatically resolves ``subscribe/reply''. The user specifies \texttt{example\_data} and \texttt{example\_request} in each Node definition, which are used to infer buffer shapes and data types when initializing ring buffers and request queues. Each Node exposes a public API (defined via \texttt{\_\_api\_\_}), whose methods are automatically wrapped for serialization on the server side and deserialization on the client side, enabling transparent bidirectional communication. Because Nodes are middleware-agnostic, they can be paired with different middleware backends depending on deployment requirements. Process synchronization is managed through ready and exit events that signal when ``publish/request'' loops are ready or exited, ensuring that user logic in the main process blocks until every Node has fully initialized and that the process cleans up when it completes.

\noindent\textbf{Middlewares.} Nodes interact with Middleware through a common interface, hiding transport-level details. For network-based communication, middlewares such as Zenoh or ZeroRpc handles serialization and transport over TCP or IPC. For high-throughput local communication, shared memory middleware uses a SharedMemoryManager to allocate ring buffers and request queues in shared memory, allowing zero-copy data exchange between processes. The shared memory implementation runs the Node's loops in a separate process and communicates buffer handles back to the parent through pipes, enabling multiple processes to access the same underlying memory regions without serialization overhead. Middlewares can be interchanged depending on the user's requirements. For instance, switching to the Thread middleware can help with debugging when orchestrating many Nodes on the same computer, since running everything multi-threaded within one process simplifies stack traces, breakpoints, and exception handling compared to multi-process or networked deployments. Alternatively, for embodiments such as mobile manipulator robots that may not have a powerful onboard computer, the user may use network-based middleware to communicate between multiple machines.

\subsection{Robot Stations}
\label{sec:method_station}

We aggregate the instantiation of all nodes that make up an environment into a single station configuration file. A robot station is instantiated through a composable dataclass configuration that specifies the hardware topology of the deployment, defining the set of robots, end effectors, and sensors, e.g., $\{$\texttt{arm, gripper, arm2, gripper2, wrist\_camera, wrist\_camera2}$\}$ for a bimanual robot station. The effect is to simplify the main routine's logic in robot control loops: a context manager initializes all server Nodes with the specified middleware backend, while client Nodes are started within nested context managers that yield proxy objects for transparent communication. While the nodes internally use the specified middleware, queues, and ring buffers, the main routine interacts only with the APIs, resulting in simple, Python-based code. This pattern enables the same application logic to operate over arbitrary station configurations without modification.

\subsection{Teleoperation and Data Collection}
\label{sec:method_teleopdata}

\noindent\textbf{Teleoperation.} We design three teleoperation scripts to support data collection. The first one controls relative end-effector poses (Keyboard, Phone, Gamepad, Spacemouse), and the second maps absolute joint positions using a leader-follower setup (ALOHA~\cite{zhao_learning_2023}, GELLO~\cite{wu_gello_2024}), for either single-arm or bimanual tabletop robot arms. The third script uses wrist pose retargeting from VR headsets such as Apple Vision Pro~\cite{park2024avp} to control the upper-body of a humanoid robot, similar to~\cite{he2024learning, he2025omnih2o}. These scripts do not assume any particular hardware platform, so teleoperation devices and robots can be swapped based on each user's needs. To ensure smooth teleoperation across devices and control frequencies, we include interpolators and signal-processing filters, e.g., low-pass filters. These can also be used during policy inference to mitigate the effects of noisy actions.

\noindent\textbf{Data Collection.} We define a recorder for logging robot demonstrations. To ensure consistent data collection across different hardware, we enforce standardized units: meters for world coordinates and radians for angular measurements. We adapt the RLDS-style format~\cite{ramos_rlds_2021} to aggregate multiple data streams into a unified state representation, as shown in~\Cref{lst:node_template}. To support different robot platforms, we introduce the concept of morphologies: abstract descriptions of a robot's structure, each defining its own set of state keys. Each morphology overloads the observation field of the RLDS step with its specific keys. This design standardizes state reporting across platforms, regardless of the underlying hardware.

One challenge with scaling robotic data collection is the sheer storage required to manage it, given that a typical robot demo consists of the internal state of the robot, as well as multiple camera streams (often including depth) and other relevant sensors. Additionally, depending on the specific policy learning setup, the data may need to be exported and processed differently. To this end, we use RoboDM~\cite{chen2025robo}, a toolkit that employs flexible compression schemes and streamlined file structures, to efficiently record demos in a highly compressed, lightweight format that is quick to save and load. For training or fine-tuning, we encourage users to write minimal converters so that the exported demos can directly integrate with their intended dataloader format.

\begin{figure*}[t]
\centering
\includegraphics[width=0.8\textwidth]{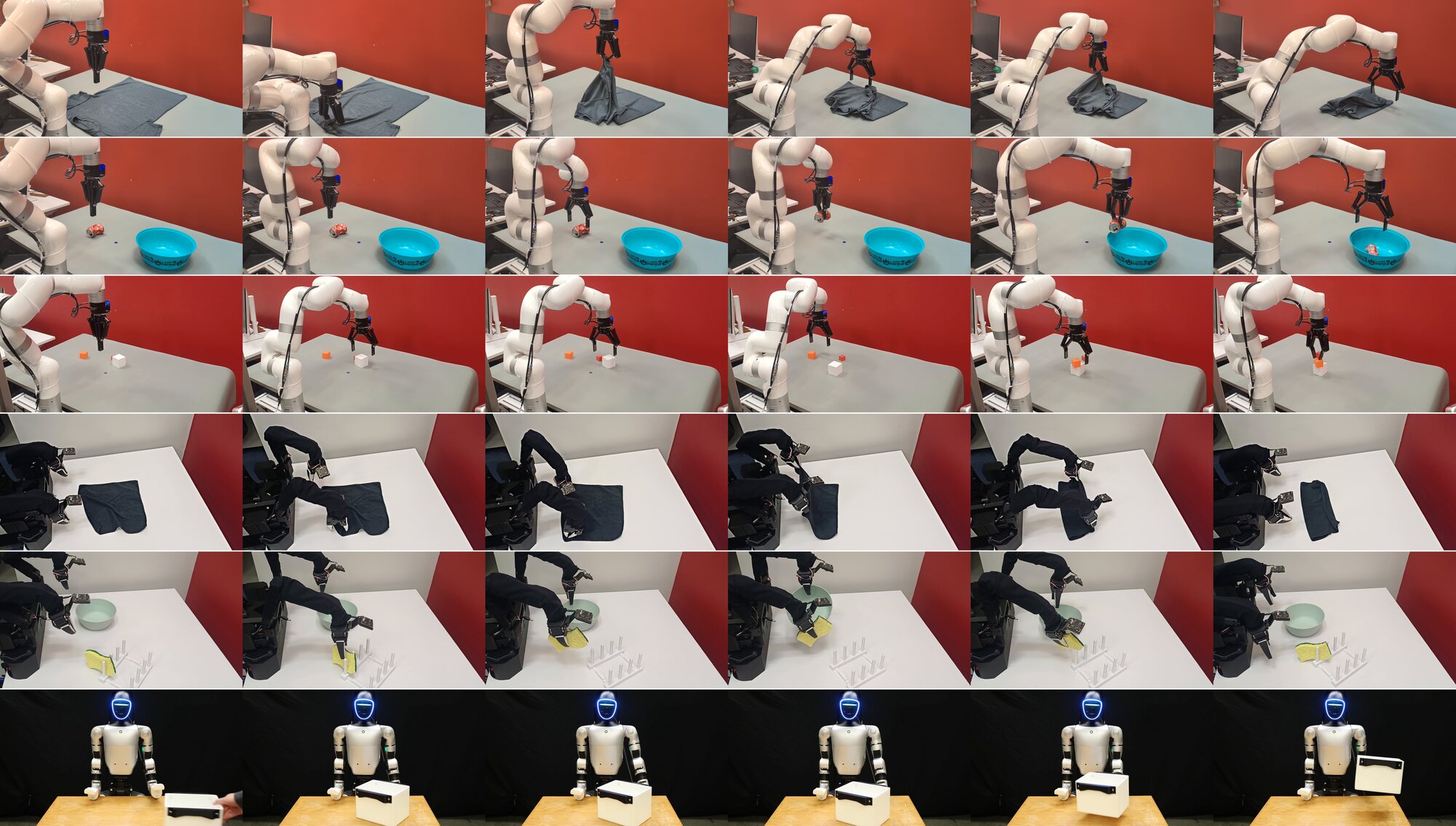}
\caption{\textbf{VLA manipulation trajectories.} We showcase rollouts of $\pi_{0.5}$ across 3 morphologies on 5 diverse tasks, shown at 0\%, 20\%, 40\%, 60\%, 80\%, and 100\% of task completion.}\vspace{-10pt}%
\label{fig:tasks}%
\end{figure*}

\begin{table*}[!ht]
\centering
{\setlength{\tabcolsep}{3pt}
\caption{\textbf{Policy deployment.} We deploy state-of-the-art policies ($\pi_{0.5}$, GR00T N1.5, Diffusion Policy) across 3 morphologies (single arm, bimanual, humanoid) and two task regimes (quasi-static and dynamic), achieving $\geq$60\% success across 20 trials on all tasks with finetuning on 50 teleoperated demonstrations. Policy rollout times closely match human demonstrations, and some tasks complete faster than demonstrations. Asynchronous inference maintains high GPU utilization throughout execution, showing that \PROJ efficiently saturates available compute. For the T-shirt folding task, half points were awarded for each fold.} %
\label{tab:policy_results}
\rowcolors{2}{white}{lightgray!15}
\begin{tabular}{l l l c c c c c c c}
\toprule
Robot & Policy & Task & \makecell{Success\\Rate} & \makecell{Task Completion\\Time (s)} & \makecell{Demo\\Time (s)} & \makecell{RAM\\(GB)} & \makecell{CPU\\(\%)} & \makecell{GPU\\Util (\%)} & \makecell{GPU\\Mem (\%)} \\
\midrule
xArm7 & BC $\pi_{0.5}$ & Fold Shirt & 92.5 & 41.96 $\pm$ \scriptsize{14.58} & 41.57 $\pm$ \scriptsize{9.25} & 22.5 $\pm$ \scriptsize{2.7} & 13.0 $\pm$ \scriptsize{1.4} & 56.7 $\pm$ \scriptsize{1.7} & 79.1 $\pm$ \scriptsize{0.0} \\
xArm7 & BC $\pi_{0.5}$ & Place Can & 95.0 & 16.08 $\pm$ \scriptsize{3.41} & 14.46 $\pm$ \scriptsize{2.00} & 24.8 $\pm$ \scriptsize{1.5} & 13.2 $\pm$ \scriptsize{1.4} & 54.6 $\pm$ \scriptsize{3.1} & 79.1 $\pm$ \scriptsize{0.1} \\
SO-100 & BC $\pi_{0.5}$  & Fold Cloth & 60.0 & 27.50 $\pm$ \scriptsize{5.51} & 22.43 $\pm$ \scriptsize{3.30} & 19.6 $\pm$ \scriptsize{0.5} & 14.1 $\pm$ \scriptsize{1.5} & 46.3 $\pm$ \scriptsize{10.0} & 78.6 $\pm$ \scriptsize{0.0} \\
SO-100 & BC $\pi_{0.5}$  & Scrub Bowl & 64.0 & 40.33 $\pm$ \scriptsize{13.68} & 27.66 $\pm$ \scriptsize{5.22} & 19.7 $\pm$ \scriptsize{0.7} & 15.0 $\pm$ \scriptsize{2.2} & 52.0 $\pm$ \scriptsize{4.8} & 78.6 $\pm$ \scriptsize{0.1} \\
Unitree G1 & BC GR00T N1.5 & Pick Box & 95.0 & 9.07 $\pm$ \scriptsize{6.10} & 10.38 $\pm$ \scriptsize{4.04} & 25.2 $\pm$ \scriptsize{0.1} & 26.6 $\pm$ \scriptsize{3.3} & 61.7 $\pm$ \scriptsize{4.7} & 33.8 $\pm$ \scriptsize{0.0} \\
\midrule
xArm7 & BC DP & Flip Tortilla & 66.7 & 12.36 $\pm$ \scriptsize{2.57} & 7.59 $\pm$ \scriptsize{0.79} & 17.2 $\pm$ \scriptsize{0.1} & 8.8 $\pm$ \scriptsize{1.9} & 8.6 $\pm$ \scriptsize{3.7} & 9.3 $\pm$ \scriptsize{0.4} \\
xArm7 & BC DP & Throw Ball & 100.0 & 14.73 $\pm$ \scriptsize{1.28} & 13.26 $\pm$ \scriptsize{2.23} & 15.8 $\pm$ \scriptsize{0.8} & 6.8 $\pm$ \scriptsize{0.1} & 0.6 $\pm$ \scriptsize{0.8} & 9.1 $\pm$ \scriptsize{0.6} \\
\midrule
Unitree G1 & RL PPO & Navigate & 100.0 & 31.27 $\pm$ \scriptsize{6.56} & N/A & 23.0 $\pm$ \scriptsize{0.1} & 10.3 $\pm$ \scriptsize{0.4} & 5.1 $\pm$ \scriptsize{0.1} & 10.3 $\pm$ \scriptsize{0.1} \\
Booster T1 & RL PPO & Navigate & 100.0 & 29.73 $\pm$ \scriptsize{4.49} & N/A & 22.6 $\pm$ \scriptsize{0.1} & 10.4 $\pm$ \scriptsize{0.4} & 5.3 $\pm$ \scriptsize{0.2} & 10.4 $\pm$ \scriptsize{0.3} \\
\bottomrule
\end{tabular}}
\end{table*}

\subsection{Policy Inference}
\label{sec:method_policy}

Aside from providing lightweight building blocks for hardware components, we also want to reduce the overhead needed to swap between different policies. To reduce the challenge of integrating robot policies into our control stack, we design a high-level API for policies. For each policy, we only require a lightweight interface to instantiate the policy, convert observations from a standardized format to the policy-specific observation format, and run inference. We design a policy wrapper node that uses this API to directly instantiate policies and asynchronously handle inference requests. By handling inference requests directly through our middleware, we avoid the additional overhead of a dedicated policy server. Additionally, we design a configurable, policy-agnostic, hardware-agnostic inference script that queries the policy wrapper, handles logic for continuously obtaining observations from hardware, post-processes actions for smoothness, and sends commands to hardware. This allows for seamless switching between different policies and hardware.

\section{Evaluation}
\label{sec:eval}

We evaluate whether \PROJ can support the robot learning workflow, from teleoperated data collection to fine-tuning and deployment, across diverse embodiments and policy classes (VLAs, Diffusion Policy, RL). Our experiments aim to answer the following questions regarding \PROJ's core functionalities:
\begin{itemize}
    \item \textit{Is \PROJ performant for real-time policy deployment?}
    \item \textit{Can \PROJ support data collection across diverse morphologies that require different teleoperation interfaces?}
    \item \textit{Does \PROJ enable effective deployment of different policy classes (VLAs, Diffusion Policy, RL) across embodiments?}
\end{itemize}
Beyond this, we also show in Appendix~\ref{app:xemb} that \PROJ readily supports cross-embodiment robot learning workflows, though effectively leveraging such multi-embodiment data to improve policy performance remains an open research problem for future exploration.

\begin{figure*}[t]
\centering
\includegraphics[width=0.8\textwidth]{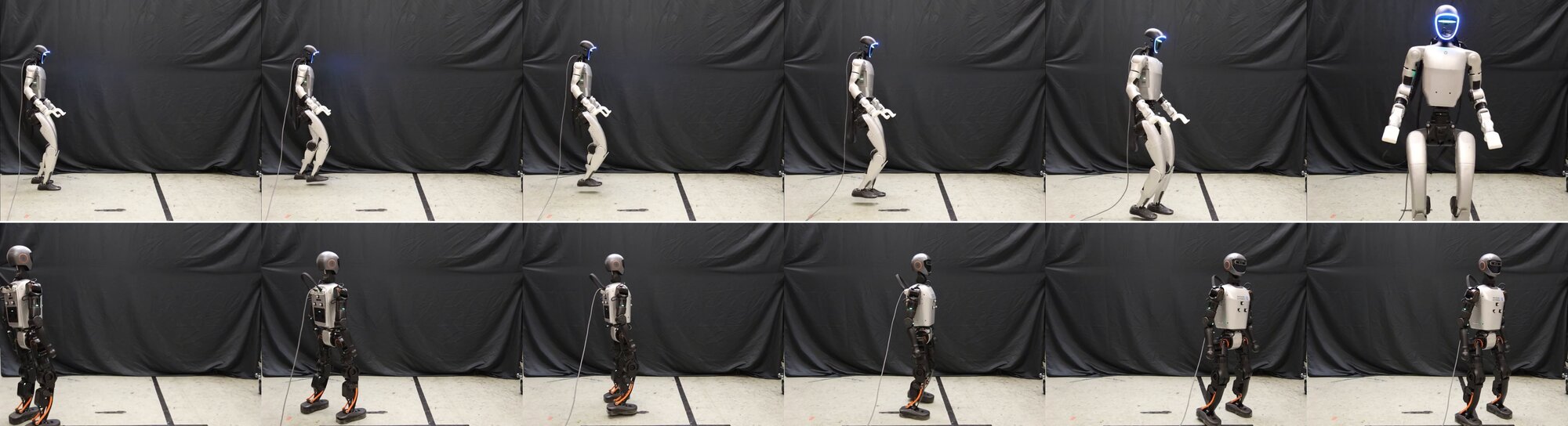}
\caption{\textbf{Humanoid locomotion trajectories.} RL policies on Unitree G1 (top) and Booster T1 (bottom), two humanoid robots from different manufacturers with different hardware drivers. \PROJ is capable of real-time control for humanoid locomotion}\vspace{-0pt}%
\label{fig:humanoid_tasks}%
\vspace{-5.0mm}%
\end{figure*}

\noindent\textbf{Experimental Setup.} All evaluations are performed on a desktop equipped with NVIDIA GeForce RTX 4090 GPU, AMD Ryzen 7 5700X CPU, and 64 GB of RAM.

\subsection{Policy Fine-tuning and Deployment}

\PROJ targets two stages of the robot learning pipeline: data collection for policy fine-tuning, and policy deployment. We adopt a bring-your-own-training-stack approach. Researchers collect demonstrations with \PROJ, export them to their preferred training format, and import the resulting policy weights back into \PROJ for deployment. This ensures consistency between data collection and policy rollout, while remaining agnostic to model architecture and training setup.

We validate this workflow across three policy families: VLAs ($\pi_{0.5}$, GR00T N1.5), Diffusion Policy (DP), and RL (PPO); spanning three morphologies (single-arm, bimanual, humanoid) and four robot platforms. The same \PROJ stack drives data collection and policy rollout for every case; only the policy and station configuration changes. 

\textbf{Training Setup.} For $\pi_{0.5}$, we fine-tune\footnote{We use one checkpoint per-task, per embodiment.} from the DROID checkpoint for single-arm tasks, and ALOHA checkpoint for bimanual tasks (20K steps each); for GR00T N1.5, we use 150 demonstrations on the humanoid manipulation task; for DP, we collect 50 demonstrations per task; locomotion policies are trained with PPO in simulation. All demonstrations are collected at 50--80\,Hz and stored in a compressed format exportable to each target training pipeline, e.g., 150 three-camera episodes occupy only 1.31\, GB. To validate policies, we report success-rate metrics and compare the average task completion time with the average demonstration time.

\noindent\textbf{Single-Arm Tabletop Manipulation.} We use a UFactory xArm7 with a Robotiq 2F-140 gripper and three RealSense D400 cameras, on two task regimes with two different policies.

\textit{Tabletop tasks with $\pi_{0.5}$.} For pick-and-place (\textit{place can}, \textit{fold shirt}), we collect demonstrations with both a Spacemouse and a GELLO leader-follower interface. Empirically, end-effector space devices such as the Spacemouse yield cleaner demonstrations for tasks where rotation is not a major factor. We also find that training VLAs with binary gripper actions yields poor gripper control, so we apply trajectory interpolation and low-pass filtering to smooth collected actions. As shown in \Cref{tab:policy_results}, both tasks achieve success rates above 90\%, with completion times within 2\,s of demonstration time.

\textit{Dynamic tasks with Diffusion Policy.} Tortilla flipping and ball throwing are both dynamic tasks that demand precise, high-frequency control, which makes them well-matched for Diffusion Policy. We collect 50 GELLO demonstrations at 80\,Hz and deploy DP through the same \PROJ pipeline. As reported in \Cref{tab:policy_results}, DP reaches 66.7\% success on \textit{flip tortilla} and 100\% on \textit{throw ball} across 20 trials, with execution times within a few seconds of the demonstration average. 

\noindent\textbf{Bimanual Tabletop Manipulation.} We use SO-100 arms to evaluate \PROJ's support for bimanual coordination with $\pi_{0.5}$. \PROJ natively supports ALOHA-style robot-to-robot teleoperation; configuring the leader arms as teleoperation devices requires only a change to the station configuration file. We perform bimanual cloth folding and bowl scrubbing. All bimanual tasks achieve at least a 60\% success rate. Bi-manual tasks exhibit longer policy completion times relative to their demonstrations than the single-arm tasks do (average difference of $-8.87$\,s), with \textit{scrub bowl} showing the largest gap. We attribute this to rollout-time retry behavior rather than system overhead: the policy often does not complete the task on the first attempt and instead reattempts until succeeding.

\noindent\textbf{Humanoids.} Beyond tabletop arms, we validate \PROJ on humanoids from two different vendors, Unitree G1 and Booster T1 (\Cref{fig:humanoid_tasks}), covering both upper-body manipulation and locomotion with an RL controller. On the Unitree G1, GR00T N1.5 reaches 95\% success on \textit{pick box} (\Cref{tab:policy_results}), with rollouts faster on average than the human demonstrations. For locomotion, RL-PPO policies trained in simulation control robots through the same \PROJ deployment path used for the manipulation policies above, achieving successful locomotion on each platform. Together, these runs show that \PROJ accommodates RL and imitation policies, vendor-specific hardware drivers, and high-frequency control required for humanoids.

\begin{figure}[h]
    \centering
    \includegraphics[width=0.95\columnwidth]{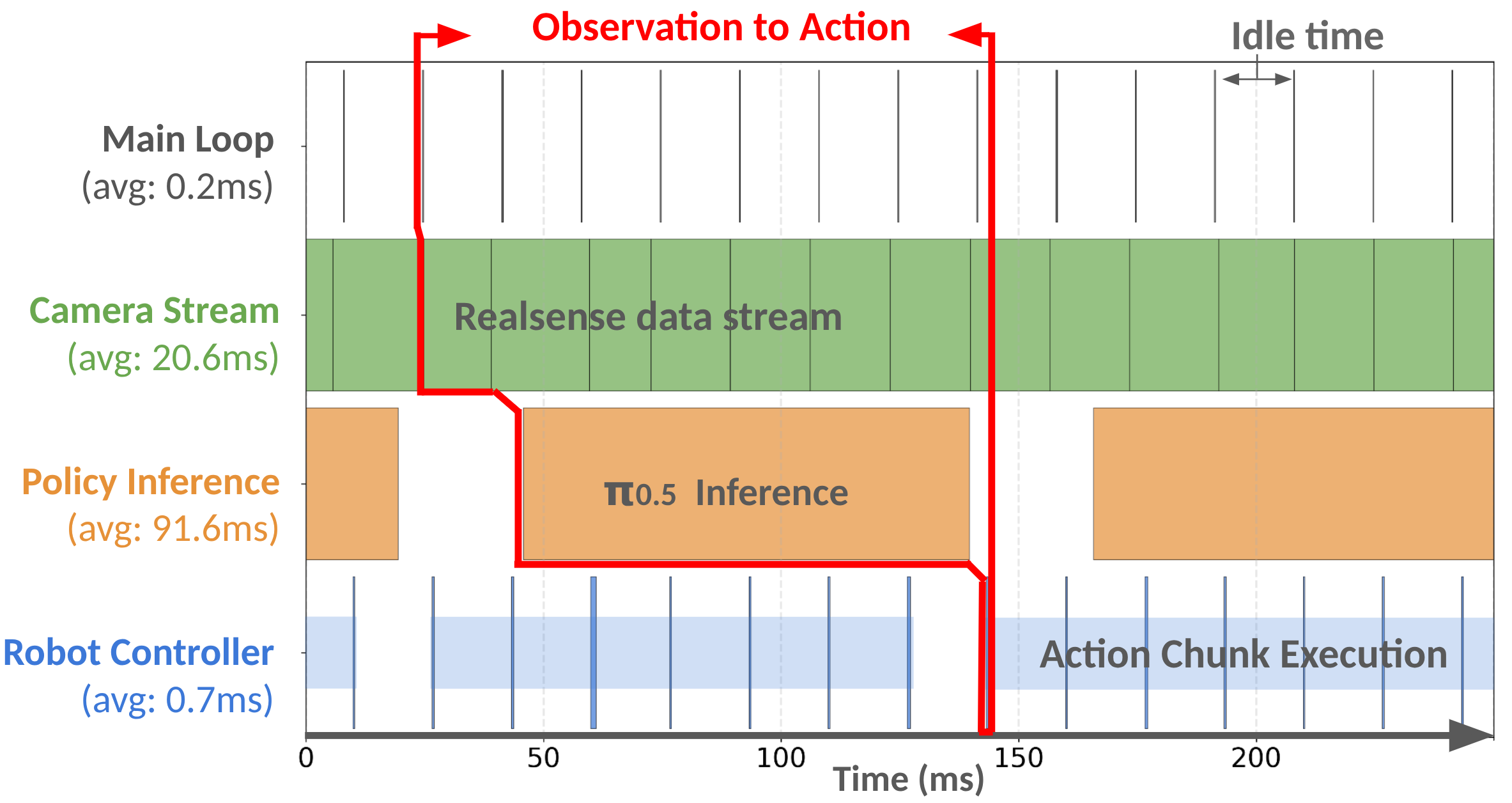}
    \caption{\textbf{Node profiling during policy deployment.} \PROJ distributes blocking operations (camera streaming, policy inference, robot control) across separate nodes, keeping the main loop free for precise timekeeping.}\vspace{-0pt}%
    \label{fig:rollout_timings}%
    \vspace{-1.0em}
\end{figure}

\subsection{Performance Analysis}

We evaluate \PROJ at two levels: middleware latency and end-to-end latency under realistic workload conditions.

\noindent\textbf{Middleware Latency.} To establish an approximate lower bound on \PROJ's communication overhead, we measure latency across all supported middlewares. We follow the Open Messaging Benchmark by defining latency as half the median round-trip time, removing the 1st and 99th percentiles to improve robustness to outliers. We use a synthetic 2048-byte payload to reflect typical observation sizes. This benchmark isolates middleware performance because the main loop performs only ring-buffer reads and writes. \Cref{tab:latency} shows results across five of \PROJ's supported backends. Zenoh and Shared Memory achieve sub-millisecond round-trip latency, making them the preferred choice for high-frequency control; Thread and ZeroRpc backends remain close at ${\sim}1$\,ms, trading a small latency penalty for easier debugging or network flexibility, while Portal adds further overhead (${\sim}2$\,ms) in exchange for its distributed-deployment features.

\vspace{-1.0em}
\begin{table}[H]
\centering
\caption{\textbf{Middleware latency.} We report latency across RIO's supported middlewares, mean $\pm$ stddev, averaged across 1,000 passes with 2048 bytes payload. Network-based backends (Zenoh, ZeroRpc, Portal) enable distributed multi-machine deployments, while local options (Shared Memory, Threads) minimize overhead for single-machine setups. This flexibility allows balancing performance, distribution, and system complexity depending on requirements.}
\rowcolors{2}{white}{lightgray!15}
\begin{tabular}{lc}
\toprule
Middleware & Latency (ms) \\
\midrule
Zenoh & 0.4287 $\pm$ 0.1344 \\
Shared Memory & 0.5413 $\pm$ 0.6166 \\
Thread & 0.9877 $\pm$ 0.3001 \\
ZeroRpc & 1.0476 $\pm$ 0.1687 \\
Portal & 1.9699 $\pm$ 0.3353 \\
\bottomrule
\end{tabular}
\label{tab:latency}
\end{table}

\vspace{-0.5em}

\begin{figure}[h]
    \centering
    \includegraphics[width=0.95\columnwidth]{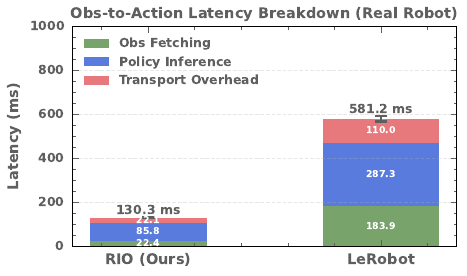}
    \caption{\textbf{Real-world observation-action latency.} End-to-end latency during $\pi_{0.5}$ deployment with an SO-100 in the loop: \PROJ reaches $130.3$\,ms versus $581.2$\,ms for LeRobot.}\vspace{-0pt}%
    \label{fig:lerobot_comparison}%
    \vspace{-0.5em}
\end{figure}

\noindent\textbf{End-to-end Profiling.} To quantify performance under realistic conditions, we profile \PROJ during $\pi_{0.5}$ rollouts, receiving inputs from three Intel RealSense cameras (two D415s and one D405) at 640 $\times$ 480 resolution. \Cref{fig:rollout_timings} shows the execution timeline across nodes: the main loop remains non-blocking, allowing for precise time-keeping, with blocking operations distributed to dedicated nodes. Asynchronous inference allows the system to preemptively request actions, maintaining continuous control despite the ${\sim}85.8$\,ms model forward pass. To compare observation-to-action latency against LeRobot, a widely adopted Python-based framework, we run action chunks from the fine-tuned $\pi_{0.5}$ policy with an SO-100 in the loop under both stacks, keeping the camera setup identical to previous tasks; the robot's own read/write overhead is negligible relative to observation capture, inference, and action execution. As shown in \Cref{fig:lerobot_comparison}, \PROJ reaches $130.3$\,ms end-to-end observation-to-action latency versus $581.2$\,ms for LeRobot. This efficiency stems from its streamlined architecture: while LeRobot threads observations before network transmission to an asynchronous policy server, \PROJ directly leverages the middleware for asynchronous inference. Lower pipeline latency directly raises the effective control frequency, which is what makes the same stack viable across policy classes from the slower VLA rollouts to the high-frequency DP and RL policies reported in \Cref{tab:policy_results}.

\begin{figure}[h]
\centering
\includegraphics[width=0.8\linewidth,trim={2cm 0 12cm 0},clip]{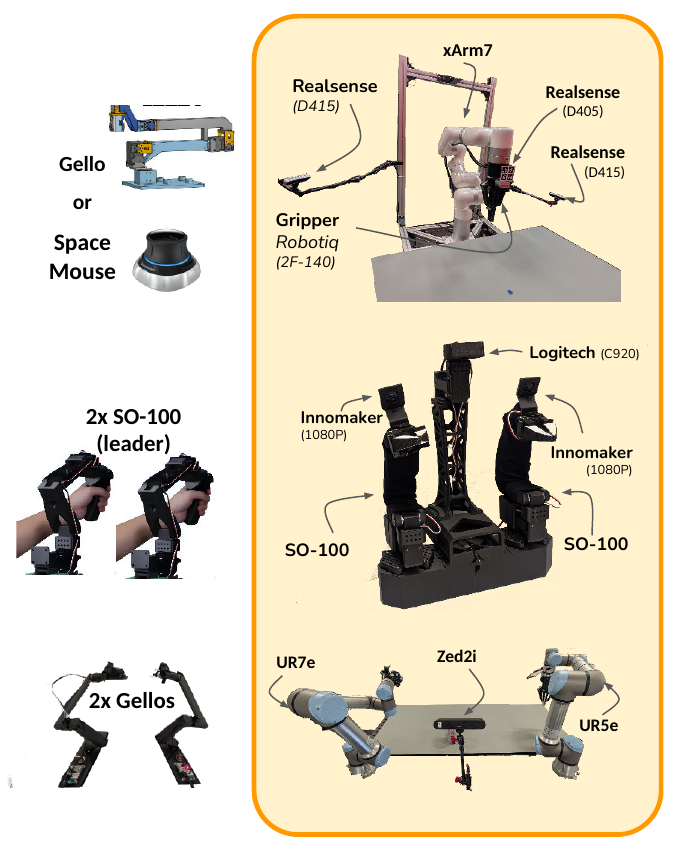}
\caption{\textbf{Example robot stations.} We illustrate single arm and bimanual robot stations with different cameras, controlled with different teleoperation interfaces using \PROJ.}
\label{fig:robot_stations}
\vspace{-1.5em}
\end{figure}

\section{Conclusion}

The lack of flexible, reusable, accessible, performant, and consistent full-stack robot infrastructure has proven to be a critical barrier to cumulative progress and collaboration in robotics. In this work, we present \PROJ, a flexible real-time \proj framework for cross-embodiment robot learning. \PROJ provides lightweight, middleware-agnostic building blocks for robot control, teleoperation, data collection, and policy deployment that can be freely combined and reconfigured across diverse hardware platforms. We validate \PROJ across the complete robot learning workflow, demonstrating its effectiveness on three morphologies (single-arm, bimanual, humanoid) and four robot platforms. We open source \PROJ, and hope that it will lower the barrier for robotics practitioners to deploy, benchmark, and iterate on modern robot learning approaches across diverse hardware configurations.

\noindent\textbf{Limitations and Future Directions.} In this work, we focus on single-embodiment fine-tuning. Since cross-embodiment generalization remains an active area of research, we primarily leave cross-embodiment fine-tuning of VLAs to future work. Additionally, while we demonstrate dynamic tasks through teleoperation, reliable dynamic task performance by VLAs remains an open research question that warrants further investigation. Future directions include systematic benchmarking of distribution shift across embodiments or extending support to include other robot hardware such as mobile manipulators and multi-fingered dexterous robot hands.

\FloatBarrier
\clearpage

\section*{Acknowledgments}
Guanya Shi holds concurrent appointments as an Assistant Professor at Carnegie Mellon University and as an Amazon Scholar. This paper describes work performed at Carnegie Mellon University and is not associated with Amazon.

\nocite{*}
\bibliographystyle{plainnat}
\bibliography{references}

\clearpage

\section*{Appendix}

\subsection{Code specifics}

\textbf{Template Node.} Our Node implementation is inspired by Diffusion Policy~\cite{chi2025diffusion} and UMI~\cite{chi2024universal}, with a main loop that publishes state through a \texttt{ring\_buffer} and processes requests received through a \texttt{request\_queue}. For \PROJ, we provide code for a template node in Figure~\ref{fig:node_template}, which users can copy from to quickly implement new Nodes, such as for a different robot or teleoperation interface. To support a range of middleware with seamless switching between them, we construct Nodes via factory functions that dynamically inherit from any middleware class that implements ``publish/request'' functionality. These factory functions can produce pairs of client and server nodes to automatically handle the ``subscribe/response'' protocol. Each middleware creates its own \texttt{ring\_buffer} and \texttt{request\_queue} based on \texttt{example\_data} and \texttt{example\_request}, along with internal functionality for message passing, that is abstracted away from the user.

\textbf{Main Loop Example.} RIO streamlines robot control development by generating matching server/client pairs from a single station configuration dataclass. The factory function introspects dataclass fields and the corresponding configurations, dynamically imports modules, and yields node factories. Servers are launched in parallel using the server manager, while clients connect through the configured middleware layer. Robot and camera nodes can be optionally aggregated into an environment class that exposes Gym-style methods \texttt{reset()}, \texttt{step()}, \texttt{get\_state()}), with the embodiment type automatically inferred from available components. Peripheral nodes not wrapped by the environment, such as teleoperation devices or visualizers, remain accessible via their keys. Within the main loop, users call node API methods directly, which internally leverage ring buffers and request queues for asynchronous interprocess communication. This architecture decouples timing constraints: servers publish sensor data and process commands at their native frequencies, while the control loop samples and issues commands at its own rate without blocking. The consistent pattern across applications enables rapid prototyping of teleoperation, policy deployment, and data collection workflows.

\begin{figure}[h!]
\begin{minted}{python}
from rio_hw import time
from rio_hw.middleware import ServerManager

from rio.envs.env import make_env
from rio.envs.poll import Interface, TeleopMode

# Factory function to create servers, clients, and environment based on configuration (args)
servers, clients, env = make_env(args)

freq = args.freq
dt = 1.0 / freq

# Starts the servers with the desired middleware
with ServerManager(args.mw, list(servers.values())):
    # Start clients
    with env, clients["teleop"]() as teleop:
        arm_target_pose = env.robot.arm.get_state()["eef_pose"].copy()
        teleop_mode = TeleopMode.TRANSLATION_ROTATION
        t_last_mode_change = time.now()
        last_gripper_cmd = 0.0

        t_start = time.now()
        env.set_start_time(t_start)
        it = 0

        while True:
            t_cycle_end = t_start + (it + 1) * dt
            t_sample = t_cycle_end - dt / 2
            time.precise_wait(t_sample)

            # Query client APIs, all non-blocking
            delta_pose, gripper_pos, t_last_mode_change, teleop_mode = Interface.poll(
                args.teleop, teleop, t_sample, t_last_mode_change, teleop_mode
            )
            if gripper_pos is not None:
                last_gripper_cmd = gripper_pos

            arm_target_pose = env.robot.make_teleop_eef_cmd(
                freq,
                teleop_mode,
                delta_pose,
                arm_target_pose,
                args.arm_cfg.max_pos_speed,
                args.arm_cfg.max_rot_speed,
            )
            action = env.robot.build_action(arm_target_pose, gripper_cmd=last_gripper_cmd)
            env.step(action, t_cmd_target=t_cycle_end + dt)

            time.precise_wait(t_cycle_end)
            it += 1
\end{minted}
\caption{\textbf{Example of a main loop.} Factory functions instantiate environments and custom clients from a single configuration file. Dynamic Inheritance forwards each component to the chosen middleware; once servers and clients are initialized, method calls pass through the storage structures (queues and ring buffers), avoiding blocking operations in the main loop.}
\label{fig:main_loop_example}
\end{figure}

\textbf{Embodiment Abstraction and State Reporting.} RIO introduces an embodiment abstraction layer that aggregates hardware-specific clients into coherent robot morphologies. The base Embodiment class defines a common interface with methods for state retrieval, command execution, and action parsing. Concrete implementations such as SingleArm combine an arm client with an optional gripper client, while Bimanual pairs two arms with their respective end-effectors. During environment initialization, the factory function introspects each embodiment class's constructor signature and automatically matches against available clients from the station configuration. This design enables seamless transitions between different robot setups, from a single-arm xArm to a bimanual SO-100 configuration, without modifying user logic.

\begin{figure}[h!]
\begin{minted}{python}
@dataclass
class Camera:
    rgb: np.ndarray | None = None
    depth: np.ndarray | None = None
    meta: dict = field(default_factory=dict)

@dataclass
class Observation:
    proprio: np.ndarray  # Defaults to policy action space
    cameras: dict[str, Camera] = field(default_factory=dict)

@dataclass
class Step:
    timestep: int | None
    observation: Observation
    instruction: str | None
    action: np.ndarray | None
    meta: dict | None = field(default_factory=dict)
\end{minted}
\caption{\textbf{Base observation schema.} Standardized state reporting across different client instances and embodiments.}
\label{lst:node_template}
\end{figure}

\begin{figure}[h!]
\begin{minted}{python}
from ..schema import Observation

@dataclass
class BimanualObs(Observation):
    # Left arm (arm1)
    arm1_proprio_eef: np.ndarray | None = None
    arm1_proprio_joints: np.ndarray | None = None
    gripper1_position: float | None = None
    hand1_pose: np.ndarray | None = None
    hand1_joints: np.ndarray | None = None

    # Right arm (arm2)
    arm2_proprio_eef: np.ndarray | None = None
    arm2_proprio_joints: np.ndarray | None = None
    gripper2_position: float | None = None
    hand2_pose: np.ndarray | None = None
    hand2_joints: np.ndarray | None = None



@dataclass
class SingleArmObs(Observation):
    proprio_eef: np.ndarray | None = None
    proprio_joints: np.ndarray | None = None
    gripper_position: float | None = None

    hand_pose: np.ndarray | None = None
    hand_joints: np.ndarray | None = None
\end{minted}
\caption{\textbf{Example of observation schema.} Morphology-specific schemas extend the base format, enabling standardized state reporting across different robot configurations.}
\label{fig:schema_examples}
\vspace{-2.0em}
\end{figure}

Each embodiment defines a dedicated observation structure that extends a common base schema, ensuring standardized data representation across morphologies. The embodiment queries all component states and camera data, returning a structured observation object, which is then wrapped into a step structure containing the timestep, instruction, observation, action, and metadata fields. This unified schema provides a consistent interface for downstream consumers such as policy networks, data recorders, and visualization tools, regardless of the underlying hardware configuration.

\begin{figure}[h]
\begin{minted}{python}
import numpy as np
from .. import time
from ..middleware import ClientFactory, ServerFactory
from ..node import Node


class Template(Node):
    __api__ = ["get_state", "send_req"]
    __pub__ = True
    __req__ = True

    def __init__(self, dtype=np.float32, *, freq: int = 100, **kwargs):
        self.dtype = dtype
        super().__init__(freq=freq, **kwargs)

    def __post_init__(self):
        self.example_data = {
            "state": np.array(..., dtype=self.dtype),
            "timestamp": time.now()}
        self.example_request = {"value": np.array(..., dtype=self.dtype)}
        self.run = self.pubreq
        super().__post_init__()

    def pubreq(self):
        rate = time.Rate(self.freq)
        self.pub_ready_event.set()
        self.req_ready_event.set()

        while not self.exit_event.is_set():
            # Publish state
            data = {"state": 0.0, "timestamp": time.now()}
            self.ring_buffer.put(data)

            # Fetch requests
            reqs = self.request_queue.get_all()
            for req in reqs:
                # Handle request...

            rate.precise_sleep()

    def get_state(self, k=None, out=None):
        return (
            self.ring_buffer.get(out=out)
            if k is None
            else self.ring_buffer.get_last_k(k=k, out=out)
        )

    def send_req(self, value):
        self.request_queue.put({"value": value})

def TemplateServer(mw, *args, **kwargs):
    return ServerFactory(mw, Template, *args, **kwargs)


def TemplateClient(mw, *args, **kwargs):
    return ClientFactory(mw, Template, *args, **kwargs)
\end{minted}
\caption{\textbf{Template node.} Nodes are constructed with a factory function by dynamic inheritance from any middleware class that implements publish/request functionality, allowing for seamless switching between different middlewares. Paired client-server nodes automatically handle subscribe/response.}
\label{fig:node_template}
\end{figure}

\subsection{Cross-Embodiment Fine-Tuning}
\label{app:xemb}

To validate that \PROJ supports a full cross-embodiment learning workflow, we fine-tune $\pi_{0.5}$ on a mixed dataset spanning two morphologies (unimanual, bimanual) and two platforms (xArm7, SO-100), collected entirely through \PROJ. Results are reported in \Cref{tab:x_embodiment}. Because \PROJ exports data in a standardized, schema-aligned format, mixing demonstrations across embodiments and deploying the resulting cross-embodiment policy required no additional tooling beyond the scripts already used for the single-embodiment experiments in Section~\ref{sec:eval}. As expected, performance drops relative to single-embodiment fine-tuning, which reflects an open research challenge rather than a framework limitation; what we demonstrate here is that the workflow itself (data collection, mixing, and rollout) is supported end-to-end through a single interface. \PROJ is also paradigm-agnostic beyond VLAs: the Diffusion Policy rollouts on dynamic tasks and the PPO-based humanoid locomotion reported in \Cref{tab:policy_results} use the same policy deployment logic.

\begin{table}[h]
  \centering
  \setlength{\tabcolsep}{3pt}
  \caption{Cross-embodiment fine-tuned policy deployment on folding tasks (single checkpoint, 20 trials each).}
  \vspace{0.25em}
  \label{tab:x_embodiment}
  \rowcolors{2}{white}{lightgray!15}
  \resizebox{\columnwidth}{!}{
\begin{tabular}{@{}ccc@{}}
\toprule
\multicolumn{1}{l}{} & Sucess Rate (\%) & Task Completion Time(s)   \\ \midrule
Bimanual (SO-100)     & 60.0               & 31.27\\
Unimanual (xArm7)    & 70.0               & 41.93                     \\ \bottomrule
\end{tabular}
  }
\end{table}

\subsection{Onboarding a New Robot with a Coding Agent}

To quantify the effort required to onboard a previously unsupported robot, we integrate the AgileX PiPER arm into \PROJ using a coding agent (Claude Code, Opus 4.6) given the \PROJ codebase and the AgileX SDK as context. We emphasize that \PROJ itself was \textbf{not} built with coding agents; this experiment only measures integration effort under a controlled procedure.

\begin{figure}[h]
\centering
\includegraphics[width=0.9\linewidth]{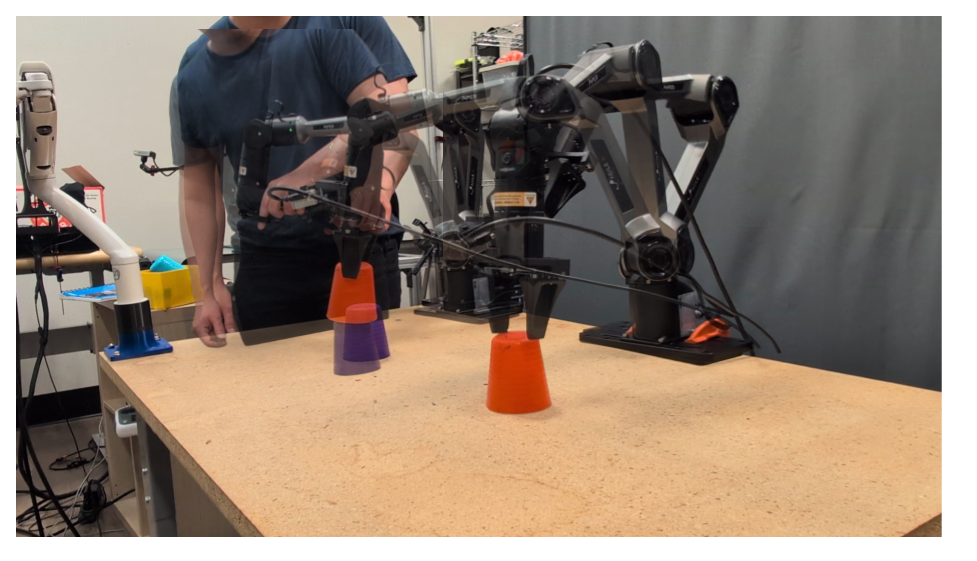}
\caption{\textbf{PiPER teleoperation (newly onboarded robot).} Cup-stacking rollout used to validate the agent-generated driver, configuration, and registry entry. Main loop and teleoperation script are unchanged from experiments in Section~IV.}
\label{fig:piper_teleop}
\end{figure}

End-to-end integration takes under one hour. Of this, roughly 40 minutes are spent on physical hardware setup (CAN bus wiring, mounting), which is unavoidable regardless of the framework. The software implementation (adding a robot driver, a station configuration, and a module-registry entry) requires \textit{9.2 minutes} and \textit{420 lines of code} across 3 files, with zero human intervention during code generation. We validate the integration by teleoperating a cup-stacking task (Figure~\ref{fig:piper_teleop}). Crucially, teleoperation runs the \textit{same} Python entry point as every other robot in our experiments; only the station configuration differs.

\end{document}